\documentclass[sigconf, screen]{acmart}
\AtBeginDocument{%
  }

\copyrightyear{2023}
\acmYear{2023}
\setcopyright{acmlicensed}
\acmConference[MM '23]{Proceedings of the 31st ACM International Conference on Multimedia}{October 29-November 3, 2023}{Ottawa, ON, Canada}
\acmBooktitle{Proceedings of the 31st ACM International Conference on Multimedia (MM '23), October 29-November 3, 2023, Ottawa, ON, Canada}
\acmPrice{15.00}
\acmDOI{10.1145/3581783.3612106}
\acmISBN{979-8-4007-0108-5/23/10}

\renewcommand\footnotetextcopyrightpermission[1]{}

\usepackage{enumitem} 
\usepackage{multirow} 
\usepackage{subcaption}
\usepackage{colortbl}  
\usepackage{bbding}
\usepackage{ulem}
\usepackage{balance}
\usepackage{breakurl}
\begin{document}

\title[Regress Before Construct]{Regress Before Construct: Regress Autoencoder for Point Cloud Self-supervised Learning}

\author{Yang Liu}
\affiliation{%
  \institution{College of Computer Science, \\Sichuan University}
  \country{China}
  }

\author{Chen Chen}
\affiliation{%
  \institution{Center for Research in Computer Vision, University of Central Florida}
  \country{USA}
}

\author{Can Wang}
\affiliation{%
 \institution{\mbox{Laboratory on Multimedia Information} \\Processing at the Department of Computer Science, Kiel University\\Hangzhou Linxrobot Company}
   \country{China}
}
\author{Xulin King}
\affiliation{%
  \institution{\mbox{Hangzhou GOTHEN Technology Co., Ltd}}
  \country{China}
}
\author{Mengyuan Liu}
\authornote{Corresponding author: liumengyuan@pku.edu.cn}
\affiliation{%
  \institution{Key Laboratory of Machine Perception, Shenzhen Graduate School, Peking University}
  \country{China}
  }
\renewcommand{\shortauthors}{Yang Liu, Chen Chen, Can Wang, Xulin King, \& Mengyuan Liu}

\begin{abstract}
Masked Autoencoders (MAE) have demonstrated promising performance in self-supervised learning for both 2D and 3D computer vision. Nevertheless, existing MAE-based methods still have certain drawbacks. Firstly, the functional decoupling between the encoder and decoder is incomplete, which limits the encoder's representation learning ability. Secondly, downstream tasks solely utilize the encoder, failing to fully leverage the knowledge acquired through the encoder-decoder architecture in the pre-text task.
In this paper, we propose Point Regress AutoEncoder (Point-RAE), a new scheme for regressive autoencoders for point cloud self-supervised learning. 
The proposed method decouples functions between the decoder and the encoder by introducing a mask regressor, which predicts the masked patch representation from the visible patch representation encoded by the encoder and the decoder reconstructs the target from the predicted masked patch representation.
By doing so, we minimize the impact of decoder updates on the representation space of the encoder.
Moreover, we introduce an alignment constraint to ensure that the representations for masked patches, predicted from the encoded representations of visible patches, are aligned with the masked patch presentations computed from the encoder.
To make full use of the knowledge learned in the pre-training stage, we design a new finetune mode for the proposed Point-RAE. Extensive experiments demonstrate that our approach is efficient during pre-training and generalizes well on various downstream tasks. Specifically, our pre-trained models achieve a high accuracy of \textbf{90.28\%} on the ScanObjectNN hardest split and \textbf{94.1\%} accuracy on ModelNet40, surpassing all the other self-supervised learning methods. 
Our code and pretrained model are public available at:  \url{https://github.com/liuyyy111/Point-RAE}.
\end{abstract}




\maketitle

\begin{figure}[t]
  \centering
  \includegraphics[width=\linewidth]{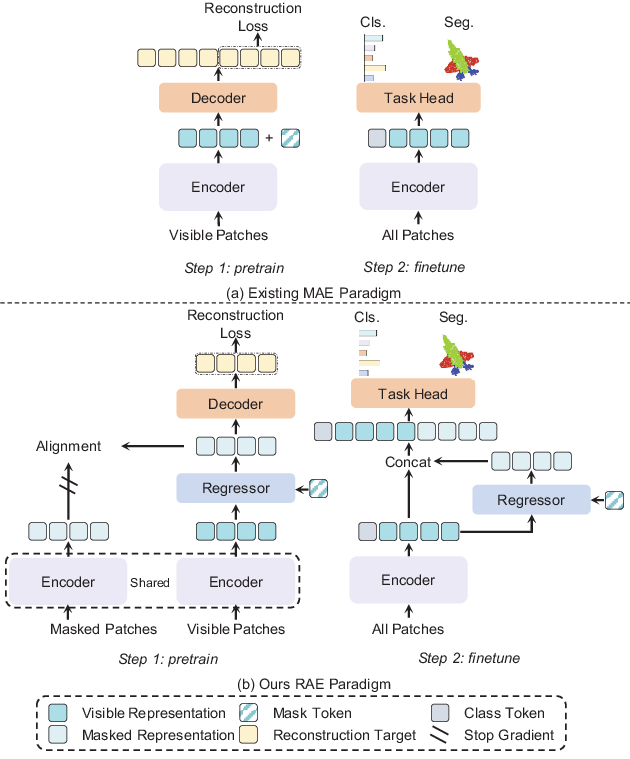}
  \caption{Differences between existing MAE-based methods (a) and our Point-RAE (b).
  During pre-training, Point-RAE predicts the representation of masked patches and uses it to reconstruct the point cloud. Decoupling the encoder and decoder through the mask regressor enables the decoder to interact with the predicted representation space, avoiding limitations on the encoder's representation capability. In fine-tuning, both the encoder and mask regressor is optimized to leverage the pre-training knowledge for downstream tasks
  }
  \label{intro}
\end{figure}
\section{Introduction}
Self-supervised learning has emerged as a prominent approach for learning representations from unlabeled data, exhibiting exceptional performance across various domains, including natural language processing \cite{devlin2018bert, brown2020language, radford2018improving, radford2019language}, computer vision \cite{he2022masked,he2020momentum,chen2020simple,chen2021exploring}, and multi-modality learning \cite{jia2021scaling, radford2021learning, zhang2022pointclip}. 
By leveraging large-scale unlabeled data for pre-training, models are equipped with robust and versatile representation capabilities, enabling them to offer substantial improvements to downstream tasks through fine-tuning.

Inspired by the great success of BERT \cite{devlin2018bert} in natural language processing (NLP) tasks and MAE \cite{he2022masked} in computer vision (CV), masked point modeling (MPM) has been introduced for 3D point cloud pre-training as a new pretext task, which randomly masking some patches of a point cloud and learning to reconstruct the masked patches. 
As pioneer work, Point-MAE \cite{pang2022masked}, Point-M2AE \cite{zhang2022point} propose to perform MPM in self-supervised pre-train with transformer \cite{vaswani2017attention}. 
They utilize asymmetric encoder-decoder transformers to apply masked autoencoding for self-supervised learning on 3D point cloud. 
Specifically, they represent the input point cloud as multiple local patches and randomly mask them with a high ratio to build the pretext task for reconstruction.
The encoder aims at capturing high-level latent representations from limited visible patches, and the lightweight decoder is focused to reconstruct masked point patches in coordinate space, then the pre-trained encoder is used to finetune the downstream task. 


Despite its effectiveness, the "encoder-decoder" architecture still suffers two main shortcomings. (1) \textbf{The functional decoupling between the encoder and decoder is incomplete, which results in a limitation of the encoder's representation learning ability.} The pre-text task aims to reconstruct masked point patches through visible point patches. However, since the encoding representations of the encoder are fed into the decoder as input, the decoder will also optimize the encoding representations during pre-training. Therefore, although the representation quality extracted by the encoder is not good enough, the decoder will also optimize this part.
(2) \textbf{Downstream tasks solely utilize the encoder and fail to fully leverage the knowledge acquired through the encoder-decoder architecture in the pre-text task.} For instance, masked tokens in the discarded decoder are learnable parameters that can predict masked point patches through visible point patches and further reconstruct them through the decoder. Consequently, the masked token is capable of comprehending the global context of the entire point cloud, but the existing architecture fails to exploit such knowledge efficiently.

As depicted in Figure \ref{intro} (b), we propose a new pre-training architecture, Point-RAE, to overcome the limitations mentioned above. The Point-RAE design decouples the encoder and decoder by introducing a mask regressor, which predicts a masked patch representation from the visible patch representation. The predicted masked patch representation is constrained to align with the masked patch representation computed by the encoder. The decoder then reconstructs the predicted masked patch representation. Furthermore, the Point-RAE design expects the encoder to take the responsibility of representation learning through two approaches: first, the latent representations of visible blocks are not updated in other parts; second, the alignment constraints expect the representations predicted by the mask regressor to also lie in the encoded representation space.
By doing so, the decoder does not directly update the representation space of the encoder but directly interacts with the predicted representation space of the mask regressor, \textbf{avoiding the limitation on the representation capability of the encoder.}
Moreover, the alignment constraint can ensure that the prediction representation space of the mask regressor is consistent with the representation space of the encoder, and it also ensures that the reconstruction function of the decoder is not affected.

We present an additional method to augment the performance of our pre-trained model. 
In the fine-tuning stage, We encode all patches in the encoder and design different architectures for different tasks to exploit the predictive power of mask regressors.
\textbf{This methodology maximizes the utilization of pre-trained knowledge and fully exploits the unique features of sparse point clouds.}
Consequently, our approach attains state-of-the-art performance on 3D downstream tasks. For instance, Point-RAE achieves a classification accuracy of \textbf{90.28\%} on the ScanObjectNN \cite{uy2019revisiting} hardest split and a classification accuracy of \textbf{94.1\%} on ModelNet40.

Our contributions are summarized as follows:
\begin{enumerate}[leftmargin=1em,parsep=1pt]
\item[$\bullet$] We introduce a novel architecture, Point-RAE, for self-supervised learning on point clouds. 
Point-RAE introduces the mask regressor to predict the masked patches before the reconstruction task, thereby reducing the direct interaction between the decoder and the encoder representation space, avoiding the limitation on the representation capability of the encoder.
\item[$\bullet$] We propose a new fine-tuning paradigm for Point-RAE, which extends beyond merely fine-tuning the encoder but uses both the encoder and mask regressor in the pre-trained architecture for downstream tasks.
Making full use of the predictive ability of the mask regressor can predict more representations of points that do not exist in the original input point cloud, and make up for the sparsity of point cloud data.
\item[$\bullet$] 
Extensive experimental results by transferring the learned representations to various benchmarks demonstrate the superiority of our proposed Point-RAE compared to recent state-of-the-art self-supervised 3D learning methods. For example, achieving 90.28\% accuracy on the most challenging PB-T50-RS benchmark and 94.1\% accuracy on ModelNet40.

\end{enumerate}

\section{Related Work}

\subsection{Transformers in Point Clouds.} 
Transformers were initially proposed to model long-term dependencies in natural language processing (NLP) tasks \cite{vaswani2017attention}, and have since achieved remarkable success in this area \cite{devlin2018bert, radford2018improving} as well as in other domains such as image and video understanding tasks \cite{dosovitskiy2020image, jiang2021transgan, radford2021learning, steiner2021train, wang2021max}. More recently, there have been efforts to apply Transformers to 3D point cloud data, with PCT \cite{guo2021pct} and Point Transformer \cite{zhao2021point} proposing novel attention mechanisms for point cloud feature aggregation, and 3DETR \cite{misra2021end} utilizing Transformer blocks and the parallel decoding strategy from DETR \cite{carion2020end} for 3D object detection.
However, utilizing the end-to-end standard Transformer architecture alone for 3D shape classification has resulted in lower performance compared to state-of-the-art methods that use point-based \cite{ma2022rethinking} and convolution-based \cite{qiu2021geometric} approaches.

\subsection{3D Point Cloud Pre-training}
Supervised learning for point clouds has achieved significant progress with delicately designed architectures \cite{wang2019dynamic,qi2017pointnet++,qi2017pointnet,guo2021pct} and local operators \cite{xu2021image2point,ma2022rethinking, li2018pointcnn}. 
However, they are confined to limited data domains \cite{wu20153d, uy2019revisiting} that they are trained on, and lack satisfactory generalization ability.
In contrast, self-supervised pre-training via unlabelled point clouds \cite{zhang2021self} has shown promising transferable ability, providing a good network initialization for downstream fine-tuning. Mainstream 3D self-supervised approaches employ encoder-decoder architectures to recover input point clouds from transformed representations, including point rearrangement \cite{sauder2019self}, part occlusion \cite{wang2021unsupervised}, rotation \cite{poursaeed2020self}, downsampling \cite{li2019pu}, and codeword encoding \cite{yang2018foldingnet}. Concurrent works also adopt contrastive pre-text tasks between 3D data pairs, such as local-global relations \cite{fu2022distillation, rao2020global, afham2022crosspoint}, temporal frames \cite{huang2021spatio}, and augmented viewpoints \cite{xie2020pointcontrast}. More recent works leverage pre-trained CLIP \cite{radford2021learning} for zero-shot 3D recognition \cite{zhu2022pointclip, zhang2022pointclip,xu2021image2point}, or introduce masked point modeling \cite{pang2022masked, zhang2022point, yu2022point} as strong 3D self-supervised learners.

\subsection{Masked Autoencoders}
Recently, masked autoencoders have become one of the hottest research directions and have shown excellent performance in both NLP \cite{devlin2018bert, liu2019roberta} and CV \cite{xie2022simmim, bao2021beit, chen2022context}. 
Motivated by BERT \cite{devlin2018bert} for masked language modeling and BEiT \cite{bao2021beit} for masked image modeling, 
Point-BERT \cite{yu2022point} uses the Transformer structure to solve the masked point modeling task, in which the aim of the pre-train task is to predict the discrete tokens.  But they do not have explicitly an encoder or a decoder, limiting the representation learning quality. Point-MAE \cite{pang2022masked}  prepend an additional lightweight Transformer structure as a decoder, the encoder only receives visible patches, while the decoder decodes the encoded representations and mask tokens to predict the masked point cloud. Point-M2AE \cite{zhang2022point} modifies the standard Transformer structure into a pyramid structure and designs a multi-scale masking strategy to incrementally model the spatial geometry and capture the fine-grained and high-level semantics of 3D shapes. 
In addition, some recent approaches \cite{dong2022autoencoders, qi2023contrast, zhang2022learning} have introduced cross-modalities such as images and text to enhance the pre-training of Masked Point Modeling tasks.
There are also some methods \cite{zhang2022masked, liu2022masked} that improve the objective of the improved Masked Point Modeling task. 

\begin{figure*}[!t]
  \centering
  \includegraphics[width=0.85\linewidth]{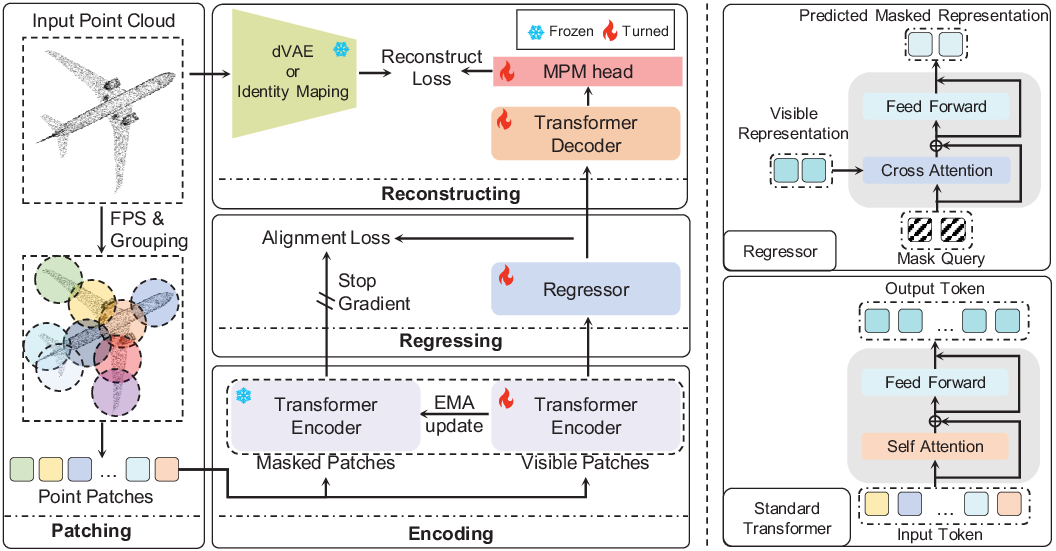}
  \caption{The pre-train pipeline of our proposed Point-RAE.
  (Left) The training process of pre-training stage consisted of patching, encoding, regressing, and reconstructing. 
  Patching: transfer the point cloud into point patches.
Encoding: encode the visible point patches to get visible representation.
Regressing: predict the representation of masked point patches.
Reconstructing: reconstruct the target by decoding the predicted representation of masked point patches.
  (Right) The difference between regressor and standard Transformer.}
  \label{framework}
\end{figure*}

\section{Preliminaries: Masked Autoencoder in Point Cloud}
We begin by providing a brief overview of the masked autoencoder framework for point clouds. Several previous works, such as \cite{yu2022point, dong2022autoencoders, pang2022masked, zhang2022point}, have employed this approach to perform 3D point cloud masked autoencoding. This framework typically includes a token embedding module, an asymmetric encoder-decoder transformer, and a reconstruction head that is responsible for reconstructing either masked 3D coordinates \cite{pang2022masked, zhang2022point} or discrete tokens \cite{yu2022point, dong2022autoencoders}.
\subsection{Patch Embedding} 
Due to the quadratic complexity of the self-attention operators, direct input of point clouds into the Transformer Encoder can result in prohibitively high computational costs. To mitigate this issue, existing MAE-based methods \cite{pang2022masked, yu2022point} adopt a patch embedding strategy that converts input point clouds into 3D point patches. 

Specifically, given a raw point cloud $\mathcal{P}\in\mathbb R^{N\times3}$, MAE-based methods initially utilize Furthest Point Sampling (FPS) to sample $S$ points ${p_i}{i=1}^S$ as patch centers. Next, k-Nearest Neighbor (k-NN) is employed to gather the $k$ nearest neighbors for each patch center, resulting in a set of 3D point patches ${g_i}{i=1}^S$. These 3D point patches are then aggregated into patch embeddings $f_i \in\mathbb R^d$ using a mini-PointNet \cite{qi2017pointnet}, where $d$ denotes the feature dimension. In this way, we obtain a set of patch embeddings $\mathcal{F} \in\mathbb R^{S\times d}$ and their center coordinates $\{p_i\}_{i=1}^S$.
Each point patch embedding represents a local spatial region and interacts with long-range features with others in the subsequent transformer.

\subsection{Asymmetric Encoder-Decoder}
To build the pre-text learning targets, Existing MAE-based methods \cite{pang2022masked, dong2022autoencoders, zhang2022learning} mask the point tokens with a high ratio, e.g., 80\%,  only using the visible ones $\mathcal{F}_{v}\in \mathbb R^{S_{v}\times d}$ as input to the transformer encoder, where $S_{v}$ represents the number of visible patches. Each encoder block comprises a self-attention layer and is pre-trained to comprehend the global 3D shape based on the remaining visible parts. Following encoding, the visible representation $\mathcal{F}_{v}^e$ is concatenated with a collection of shared learnable masked tokens $\mathcal{T}_{m}\in\mathbb R^{S_{m}\times d}$ as the input of the decoder, where $S_{m}$ denotes the number of masked tokens and $S = S_{m} + S_{v}$. In the lightweight transformer decoder, the masked tokens learn to capture informative spatial cues from the visible ones, decode the masked tokens $\mathcal{T}_{m}$, and output the decoded masked tokens $\mathcal{T}_{m}^e$.
  
\subsection{Reconstruction}

To reconstruct the masked point patches or tokens, the final layer of the MAE architecture is the prediction head. This head is responsible for generating the output by mapping the learned features to the desired reconstruction target. In existing MAE-based methods \cite{pang2022masked, dong2022autoencoders}, a simple fully connected (FC) layer is used as the prediction head. Specifically, taking the output $\mathcal{T}_{m}^e$ from the decoder, the prediction head projects it to a vector $\mathcal{F}_{pre}$, which has the same number of dimensions as the reconstruction target. This process allows the model to reconstruct the masked regions based on the learned features and generate the final output.


The reconstruction target for existing MAE-based methods is to recover the coordinates \cite{pang2022masked} or the discrete token \cite{dong2022autoencoders} of the points in every masked point patch.
To calculate the reconstruction loss, the $l_2$ Chamfer Distance \cite{fan2017point} is used for reconstructing the coordinates, where the prediction point patches $\mathcal{F}_{pre}$ and ground truth $\mathcal{F}_{gt}$ are compared. Alternatively, the negative cosine similarity \cite{dong2022autoencoders} or cross-entropy loss \cite{yu2022point} is used as the reconstruction loss for reconstructing the discrete tokens, where the prediction tokens are compared with the ground truth tokens.

\section{Point Regress Autoencoder}
To address the issue of incomplete functional decoupling between the encoder and the decoder, we propose a novel architecture called the Point-Regression Autoencoder (Point-RAE), as depicted in Figure \ref{framework}.
The key idea of the Point-RAE is to predict masked patches in the encoded representation space from visible patches and then map the predicted representations of masked patches to the corresponding targets, which helps to decouple the encoder and decoder.

\subsection{Mask Regressor}
The mask regressor is a crucial component of Point-RAE, responsible for predicting the masked representations $\mathcal{F}_m^{\hat{e}}$ for the masked patches based on the visible representations $\mathcal{F}_v^e$ output from the encoder.
As depicted in Figure \ref{framework}, the mask regressor consists of a series of \textit{cross-attention} layers and feed-forward networks.
To ensure that the mask regressor can learn a robust mapping between visible and masked patches, we introduce the mask query by shifting the mask token of the decoder into the mask regressor. The mask query is a learnable parameter and is shared among all the masked patches. 
In the mask regressor, we mask tokens as queries, and the output of the previous cross-attention layer consists of keys and values\footnote{The key and the value in the first layer is the visible representation $\mathcal{F}_{v}^e$.} to compute cross-attention and predict representations for mask patches $\mathcal{F}_{m}^{\hat{e}}$.

\subsection{Alignment Constraint}
The latent representation alignment constraint is imposed on the latent representations $\mathcal{F}_m^{\hat{e}}$ of the masked patches predicted by the mask regressor. 
We feed the masked patches embedding $\mathcal{F}$ into the encoder to generate the representations $\mathcal{F}_m^e$, the encoder is the same as the one for encoding visible patches, but using an exponential moving average (EMA) to update the weights.
We then align the two latent representations $\mathcal{F}_m^e$ and $\mathcal{F}_m^{\hat{e}}$ for the masked patches.
By doing so, the predicted representations also lie in the encoded representation space and making predictions in the encoded representation space encourages that the encoded representations take on a larger extent of semantics.

\begin{table}
\caption{
 Real-world 3D Classification on ScanObjectNN \cite{uy2019revisiting}. We report the accuracy (\%) on the official three splits of ScanObjectNN. The best performances are in \textcolor{blue}{blue}.
}
\centering
\setlength{\tabcolsep}{1.1mm}
\begin{tabular}{@{}lcccc}
\toprule[1pt]
Method & Year &OBJ-BG  &  OBJ-ONLY & PB-T50-RS
\\
\midrule[0.5pt]
\multicolumn{5}{l}{
\hspace{-0.5em}\textbf{Supervised Learning Only}}
\\
PointNet \cite{qi2017pointnet} &2016&73.3& 79.2& 68.0 \\
PointNet++ \cite{qi2017pointnet++} &2017&82.3& 84.3& 77.9  \\
DGCNN \cite{wang2019dynamic} &2019&82.8& 86.2&  78.1 \\
PointCNN \cite{li2018pointcnn} &2018& 86.1 &85.5 &78.5  \\
SimpleView \cite{goyal2021revisiting}&2021 &- &- &80.5±0.3  \\
GBNet \cite{qiu2021geometric} &2021&-& -& 81.0  \\
PRA-Net \cite{cheng2021net}&2021 &-& -& 81.0  \\
MVTN \cite{hamdi2021mvtn}&2021 & 92.6 &92.3 &82.8  \\
PointMLP \cite{ma2022rethinking} &2022&-& -& 85.4±0.3  \\
PointNeXt \cite{qian2022pointnext} &2022&-& -& 87.7±0.4  \\
P2P-RN101 \cite{wang2022p2p} &2022&-& -& 87.4 \\
P2P-HorNet \cite{wang2022p2p} &2022&-& -&  89.3  \\
\midrule[0.5pt]
\multicolumn{5}{l}{
\hspace{-0.5em}\textbf{with Self-Supervised Representation Learning (\textit{FULL})}}
\\
Transformer \cite{vaswani2017attention} &2017&79.86 &80.55 &77.24 \\
Point-BERT \cite{yu2022point} &2022&87.43 &88.12 &83.07  \\
MaskPoint \cite{liu2022masked} &2022&89.30 &88.10 &84.30 \\
Point-MAE \cite{pang2022masked} &2022 &90.02 &88.29 &85.18  \\
Point-M2AE \cite{zhang2022masked} &2022&91.22 &88.81 &86.43 \\
ACT \cite{dong2022autoencoders} &2023&93.29 &91.91 &88.21 \\
I2P-MAE \cite{zhang2022learning} &2023&94.15 &91.57 &90.11 \\
\textbf{Ours:Point-RAE} & &\textcolor{blue}{\bf 95.53} &\textcolor{blue}{\bf 93.63} &\textcolor{blue}{\bf 90.28} \\
\midrule[0.5pt]
\multicolumn{5}{l}{
\hspace{-0.5em}\textbf{with Self-Supervised Representation Learning (\textit{LINEAR})}}
\\
Point-MAE \cite{pang2022masked} &2022&82.58±0.58 &83.52±0.41 &73.08±0.30\\
ACT \cite{dong2022autoencoders} &2023&85.20±0.83 &85.84±0.15 &76.31±0.26   \\
\textbf{Ours:Point-RAE} && \textcolor{blue}{\bf 86.15$\pm$0.33}& \textcolor{blue}{\bf 86.31$\pm$0.23}& \textcolor{blue}{\bf 78.25$\pm$0.30}\\
\midrule[0.5pt]
\multicolumn{5}{l}{
\hspace{-0.5em}\textbf{with Self-Supervised Representation Learning (\textit{MLP-3})}}
\\
Point-MAE \cite{pang2022masked}&2022 & 84.29±0.55 &85.24±0.67 &77.34±0.12\\
ACT \cite{dong2022autoencoders} &2023& 87.14±0.22  &88.90±0.40 &81.52±0.19
 \\
\textbf{Ours:Point-RAE} && \textcolor{blue}{\bf 88.31$\pm$0.20}& \textcolor{blue}{\bf 89.53$\pm$0.58}&\textcolor{blue}{\bf 83.01$\pm$0.15}\\
\bottomrule[1pt]

\end{tabular}
\label{scan}
\end{table}

\subsection{Encoder-Decoder With Mask Regressor}
The proposed Point-RAE architecture comprises an encoder, a mask regressor with an alignment constraint, and a decoder. The encoder is similar to previous works and is composed of a sequence of transformer blocks based on \textit{self-attention}. 
By inputting a set point patch embedding $\mathcal{F}$, we can obtain the visible representation $\mathcal{F}_{v}^e$ from the encoder. Subsequently, the visible representation $\mathcal{F}_{v}^e$ is fed into the mask regressor to predict the masked representation $\mathcal{F}_m^{\hat{e}}$. The predicted masked representation $\mathcal{F}_m^{\hat{e}}$ is constrained by alignment with the mask representations $\mathcal{F}_m^e$ computed from the encoder. A lightweight decoder is employed to decode the predicted masked representation $\mathcal{F}_m^{\hat{e}}$ to obtain the decoded mask tokens $\mathcal{T}_{m}^e$. 
Noted that the decoder in our Point-RAE architecture differs from previous works in that it is without the mask tokens and the input of the decoder is not the concatenation of the visible representation $\mathcal{F}_{v}^e$ and mask tokens $\mathcal{T}_m$, but instead the predicted mask representation $\mathcal{F}_m^{\hat{e}}$. 
Finally, the decoded mask tokens $\mathcal{T}_{m}^e$ are fed into the reconstruction head to obtain the reconstruction target $\mathcal{F}_{pre}$.

The key idea of our Point-RAE architecture is to decouple the encoder and decoder by employing a mask regressor to predict the representation of the masked patches. Unlike the standard Transformer used in the encoder and decoder, the mask regressor comprises a series of cross-attention layers. By utilizing cross-attention, the predicted mask representation can be made independent of the mask query and exist in the same representation space as the encoder output. This approach helps reduce the impact of decoder updates on the representation space of the encoder, thereby ensuring that the encoder's learned features remain stable.

\begin{table}
\caption{
 Synthetic 3D Classification on ModelNet40 \cite{wu20153d}. We
report the accuracy (\%) before and after the voting \cite{liu2019relation}. The best performances are in \textcolor{blue}{blue}.
}
\centering
\begin{tabular}{@{}lccc}
\toprule[1pt]
Method &Year& w/o voting  &  w/ voting
\\
\midrule[0.5pt]
\multicolumn{4}{l}{
\hspace{-0.5em}\textbf{Supervised Learning Only}}
\\
PointNet \cite{qi2017pointnet} &2016&89.2& 90.8 \\
PointNet++ \cite{qi2017pointnet++} &2017&90.7& 91.9\\
DGCNN \cite{wang2019dynamic} &2019&92.9& - \\
PointCNN \cite{li2018pointcnn} &2018&92.2 &-   \\
SimpleView \cite{goyal2021revisiting}&2021 &93.9 &-   \\
GBNet \cite{qiu2021geometric} &2021&93.8& -  \\
PRA-Net \cite{cheng2021net} &2021&93.7& -  \\
MVTN \cite{hamdi2021mvtn} &2021&93.8 &- \\
PointMLP \cite{ma2022rethinking}&2022 &94.5& -  \\
PointNeXt \cite{qian2022pointnext} &2022&94.0& -  \\
P2P-RN101 \cite{wang2022p2p} &2022&93.1& -\\
P2P-HorNet \cite{wang2022p2p} &2022&94.0& -  \\
\midrule[0.5pt]
\multicolumn{4}{l}{\hspace{-0.5em}\textbf{with Self-Supervised Representation Learning (FULL)}}
\\
Transformer \cite{vaswani2017attention} &2017&91.4 &91.8 \\
Point-BERT \cite{yu2022point} &2022&93.2  &93.8  \\
MaskPoint \cite{liu2022masked}&2022 &93.8 & - \\
Point-MAE \cite{pang2022masked}&2022 &93.8 &94.0\\
Point-M2AE \cite{zhang2022masked} &2022&\textcolor{blue}{94.0} &-\\
ACT \cite{dong2022autoencoders}&2023 &93.7 &94.0  \\
I2P-MAE \cite{zhang2022learning}&2023 &93.7 &\textcolor{blue}{\bf94.1}  \\
\textbf{Ours:Point-RAE}  &&\textcolor{blue}{\bf 94.0} &\textcolor{blue}{\bf 94.1} \\

\bottomrule[1pt]

\end{tabular}
\label{modelnet}
\end{table}

\subsection{Optimization Target}
\noindent \textbf{Reconstruction Target.}
In line with \cite{dong2022autoencoders}, we adopt the dVAE tokens as the reconstruction target. Specifically, given the prediction point patches $\mathcal{F}_{pre}$ and ground truth $\mathcal{F}_{gt}$, we minimizes the negative cosine similarity $\mathcal{L}_{cos}(s,t) = 1 - \frac{s \cdot t}{\lvert s \rvert\lvert t \rvert}$ to define the reconstruction loss as follows:
\begin{equation}
    \mathcal{L}_{rec} = -\sum_{i=1}^S{\mathcal{L}_{cos}(\mathcal{F}_{pre}, \mathcal{F}_{gt}})
\end{equation}

\noindent \textbf{Alignment Target.}
To align the predicted mask representation $\mathcal{F}_m^{\hat{e}}$ with the mask representation $\mathcal{F}_m^e$, lot of loss functions can be used as alignment target.
To keep the feature space consistent, we also minimize the negative cosine similarity as the alignment loss:
\begin{equation}
    \mathcal{L}_{align} = -\sum_{i=1}^S{\mathcal{L}_{cos}(
    \mathcal{F}_m^e, sg[\mathcal{F}_m^{\hat{e}}]}
    )
\end{equation}
where $sg[\cdot]$ stands for stop gradient.
We study the effect of different loss functions as alignment targets, more details are in \textit{Appendix}.

The overall pre-training loss function is defined as the sum of the reconstruction and alignment losses:
\begin{equation}
    \mathcal{L} = \mathcal{L}_{rec} + \mathcal{L}_{align}
\end{equation}

\subsection{Fine-tune With Regressor}
In the fine-tuning stage, existing MAE-based methods direct drop the decoder and only fine-tune the encoder for downstream tasks, which does not take full advantage of the information learned by pre-training, e.g., the mask token in the decoder which has the ability to percept the global structure of the point cloud.
We propose a new fine-tune paradigm for our Point-RAE, which fine-tune both encoder and mask regressor for downstream tasks.

Since the mask regressor can predict the masked point cloud representation, we can make full use of this ability in downstream tasks, and we can use the existing point cloud structure to predict new ones that do not exist in the original point cloud $\mathcal{P}$ The representation of coordinates makes up for the sparseness of point clouds. 
In Section \ref{lab:abaltion}, we study the different fine-tuning paradigms for different tasks.

\section{Experiment}
\subsection{Implementation Details}
We employed ShapeNet \cite{chang2015shapenet} as our pre-training dataset for the purposes of object classification, part segmentation, and few-shot classification. 
The dataset comprises more than 50,000 distinct 3D models from 55 commonly occurring object categories. 
Each 3D model is sampled via farthest point sampling (FPS) to obtain 1024 points for each instance. The pre-training process uses an AdamW optimizer \cite{loshchilov2017decoupled} and cosine learning rate decay \cite{loshchilov2016sgdr}, with an initial learning rate of 0.001 and a weight decay of 0.05. 
The model is trained for 300 epochs with a batch size of 128.
We refer to the {\it Appendix} for { full implementation details} and more results.


\begin{table}
\caption{
 Few-shot classification results on ModelNet40. Overall accuracy (\%) without voting is reported. The best performances are in \textcolor{blue}{blue}.
}
\centering
\setlength{\tabcolsep}{0.9mm}
\begin{tabular}{@{}lcccc}
\toprule[1pt]
\multirow{2}{*}{\textbf{Method}}& \multicolumn{2}{c}{{5-way}}& \multicolumn{2}{c}{{10-way}}\\
\cmidrule[0.5pt](lr){2-3} \cmidrule[0.5pt](lr){4-5}
& 10-shot & 20-shot & 10-shot & 20-shot\\
\midrule[0.5pt]
DGCNN \cite{wang2019dynamic} &31.6 ± 2.8 &40.8 ± 4.6 &19.9 ± 2.1 &16.9 ± 1.5 \\
DGCNN+OcCo\cite{wang2021unsupervised} & 90.6 ± 2.8 &92.5 ± 1.9 &82.9 ± 1.3 &86.5 ± 2.2 \\
Transformer \cite{vaswani2017attention} &87.8 ± 5.2 &93.3 ± 4.3 &84.6 ± 5.5 &89.4 ± 6.3\\
Point-BERT \cite{yu2022point} &94.6 ± 3.1 &96.3 ± 2.7 &91.0 ± 5.4 &92.7 ± 5.1\\
MaskPoint \cite{liu2022masked} &95.0 ± 3.7 &97.2 ± 1.7 &91.4 ± 4.0 &93.4 ± 3.5\\
Point-MAE \cite{pang2022masked} & 96.3 ± 2.5 &97.8 ± 1.8 &92.6 ± 4.1 &95.0 ± 3.0\\
Point-M2AE \cite{zhang2022masked} & 96.8 ± 1.8 &98.3 ± 1.4 &92.3 ± 4.5 &95.0 ± 3.0\\
ACT \cite{dong2022autoencoders} &96.8 ± 2.3 &98.0 ± 1.4 &93.3 ± 4.0 &95.6 ± 2.8\\
I2P-MAE \cite{zhang2022learning} &97.0 ± 1.8 &98.3 ± 1.3 &92.6 ± 5.0 &95.5 ± 3.0\\
\textbf{Ours:Point-RAE}  &\textcolor{blue}{\bf 97.3 $\pm$1.6} &\textcolor{blue}{\bf 98.7 $\pm$1.3} &\textcolor{blue}{\bf 93.3 $\pm$4.0} &\textcolor{blue}{\bf 95.8 $\pm$3.0}\\

\bottomrule[1pt]

\end{tabular}
\label{few-shot}
\end{table}

\begin{table*}[t]
    \caption{Ablation experiments for pre-train settings on the most challenging ScanObjectNN PB-T50-RS benchmark. 
    We report the accuracy (\%) of three variants of transfer learning protocols  for 3D object recognition.
    If not specified, the default is: the decoder has depth 2, the mask regressor has depth 2, the alignment loss is negative cosine similarity,
    the masking ratio is 80\%, the pre-training length is 300 epochs and only fine-tune the encoder.
    Default settings are marked in 
    \colorbox{gray!40}{gray}.
    }
    \label{table:t3}
    \begin{subtable}{0.3 \linewidth}
      \centering
        \begin{tabular}{cccc}
            Reg. Depth & FULL & LINEAR & MLP-3\\
            \midrule
             2 &89.89 & 78.16 & 82.79\\
             4 &\cellcolor{gray!40}\textbf{90.28} & \cellcolor{gray!40}78.55 & \cellcolor{gray!40}{83.16}\\
             8 &90.06 & 78.58& \textbf{83.20} \\
             12 &89.93 & \textbf{78.63}& 83.18 \\
        \end{tabular}
        \caption{Regressor Depth. \textit{A deep regressor can improve LINEAR and MLP-3 evaluation.}}
    \end{subtable} 
    \hspace{2em}
    \begin{subtable}{0.3 \linewidth}
      \centering
        \begin{tabular}{cccc}
            Dec. Depth & FULL & LINEAR & MLP-3\\
            \midrule
             0 &  89.05 & 77.81 &82.52\\
             1 &  89.98& 78.46&82.78\\
             2 & \cellcolor{gray!40}\textbf{90.28} & \cellcolor{gray!40}\textbf{78.55} & \cellcolor{gray!40}83.16\\
             4 & 90.03 &78.48 &\textbf{83.20}\\ 
        \end{tabular}
        \caption{Decoder Depth. \textit{Performance is not sensitive to the decoder depth.}}
    \end{subtable} 
     \hspace{2em}
    \begin{subtable}{.3\linewidth}
      \centering
        \begin{tabular}{ccccc}
            Dec. & Reg. & FULL & LINEAR & MLP-3\\
            \midrule
             \XSolidBrush &\XSolidBrush &88.89 & 76.31& 81.29\\
             \Checkmark &\XSolidBrush &89.42 &77.86 &82.53\\
             \XSolidBrush &\Checkmark &89.59 & 77.65& 83.01\\
             \Checkmark &\Checkmark&\cellcolor{gray!40}\textbf{90.28} &  \cellcolor{gray!40}\textbf{78.55} & \cellcolor{gray!40}\textbf{83.16}\\
        \end{tabular}
        \caption{Regress \& Construct. \textit{
        Regress and construct are critical to learning representation.
}}
    \end{subtable}%
    
    
             
    \label{ablation}
\end{table*}
\subsection{Transfer Learning on Downstream Tasks}
\noindent \textbf{Transfer Protocol.}
The study follows the transfer learning protocols for 3D object recognition tasks proposed in \cite{pang2022masked, dong2022autoencoders}, which includes three variants:
\begin{enumerate}[leftmargin=1em,parsep=1pt]
\item[${\dagger}$] \textit{FULL}: This protocol fine-tunes the pre-trained models by updating all the parameters of the backbone and classification head.
\item[${\dagger}$] \textit{LINEAR}: In this protocol, the classification head consists of a single-layer linear MLP. During fine-tuning, only the parameters of this classification head are updated.
\item[${\dagger}$] \textit{MLP-3}: In this protocol, the classification head consists of a 3-layer non-linear MLP. . During fine-tuning, only the parameters of this classification head are updated.
\end{enumerate}

\noindent \textbf{3D Real-World Object Recognition.}
ScanObjectNN \cite{uy2019revisiting} is a challenging point cloud object dataset that is created from real-world scans, comprising of 2,902 samples from 15 categories, and includes background and occlusions, which adds to its complexity. We conducted experiments on three variants of ScanObjectNN, namely OBJ-BG, OBJ-ONLY, and PB-T50-RS. During training, we used simple Rotation as data augmentation following \cite{dong2022autoencoders}, and did not employ any voting methods during testing. The results are presented in Table \ref{scan}. Our observations are as follows:
(i) Our Point-RAE model achieves a significant improvement of +13.94\% accuracy averaged on the three variant ScanObjectNN benchmarks, compared to the Transformer \textit{from scratch} baseline under the FULL tuning protocol.
(ii) Our Point-RAE outperforms other self-supervised learning methods and achieves the best generalization across all transferring protocols on ScanObjectNN. Specifically, it achieves an average accuracy improvement of bf+5.3\% over Point-MAE on the three variant ScanObjectNN benchmarks.
(iii) Our Point-RAE sets a new state-of-the-art performance on ScanObjectNN, achieving 90.28\% accuracy on the most challenging PB-T50-RS benchmark when compared to all other methods.

\noindent \textbf{3D Synthetic Object Recognition}
We conducted an evaluation of our pre-trained model for object classification on the ModelNet40 dataset \cite{wu20153d}, which comprises 12,311 clean 3D CAD models belonging to 40 object categories. 
During training, we employed standard random \textit{Scale\&Translate} for data augmentation. Moreover, we used the standard voting method \cite{liu2019relation} during testing. The experimental results are summarized in Table \ref{modelnet}, indicating that our Point-RAE method brought significant improvements of +2.6\% compared to the Transformer \textit{from scratch} baseline under FULL transferring baseline. Additionally, our results are comparable or better than other self-supervised learning methods.

\noindent \textbf{Few-shot Object Recognition.}
We conducted fine-tuning experiments on ModelNet40 \cite{wu20153d} for few-shot classification, and the results are presented in Table \ref{few-shot}. To train the model, we used the same settings and few-shot dataset splits as in previous work \cite{yu2022point, pang2022masked}.
We followed the standard protocol and performed 10 independent experiments for each setting, reporting the mean accuracy with standard deviation.
Our Point-RAE model showed significant improvements compared to the Transformer \textit{from scratch} baseline, with an increase in accuracy of +0.5\%, +5.4\%, +8.7\%, and +6.4\% for the four few-shot settings, respectively. Furthermore, our Point-RAE consistently outperformed other self-supervised learning methods in all settings.

\begin{figure}[t]
  \centering
  \includegraphics[width=0.9\linewidth]{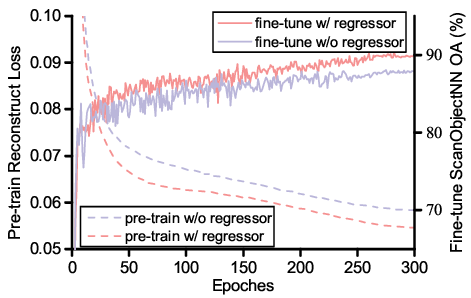}
  \caption{Pre-training reconstruct loss on ShapeNet using models w/ and w/o the mask regressor, and the fine-tuning accuracy (\%) on ScanObjectNN PB-T50-RS benchmark with only fine-tuning the encoder.
  }
  \label{loss}
\end{figure}
\subsection{Ablation Study}
\label{lab:abaltion}

\begin{figure*}[t]
  \centering
  \includegraphics[width=0.8\linewidth]{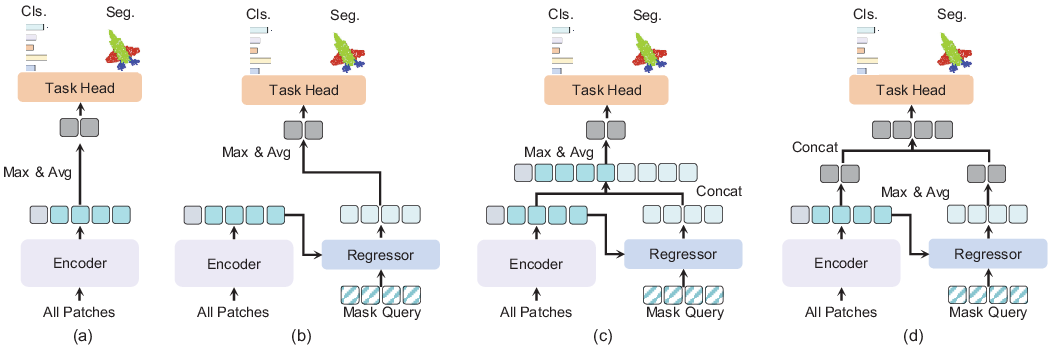}
  \caption{Ablations illustration.
(a)  the vanilla fine-tuning pipeline, which only uses the encoded representation by Mas\&Avg pooling to handle downstream tasks.
(b) proposed fine-tuning pipeline, which only uses the predicted representation by Mas\&Avg pooling to handle downstream tasks.
(c) proposed fine-tuning pipeline, which first concatenates the encoded representation and predicted representation, then feeds into Mas\&Avg pooling to handle downstream tasks.
(d) proposed fine-tuning pipeline,  which first feed encoded representation and predicted representation into Mas\&Avg pooling respectively, and concatenates the results to handle downstream tasks.}
  \label{finetune}
\end{figure*}

To verify the effectiveness of each component of our Point-RAE, we conduct ablation experiments on the settings of the pre-training stage and the fine-tuning stage respectively. We conduct an extensive ablation study on the most challenging ScanObjectNN PB-T50-RS benchmark with 2,048 input points.
In the pre-training stage, we introduced a mask regressor to decouple the encoder and decoder, avoiding the limitation of the encoder's representation learning ability. 
In the fine-tuning stage, as shown in Figure \ref{finetune}, we fully use the knowledge learned in the pre-training stage and design a new fine-tuning paradigm. 
The ablation experiments are shown in Table \ref{ablation} and Table \ref{tab:finetune} respectively.

\noindent \textbf{Regressor Depth.} 
We examine the impact of the mask regressor depth on the pre-training performance of our Point-RAE.
Table \ref{ablation} (a) varies the regressor depth (number of mask regressor blocks). 
\uline{Our analysis reveals that a sufficiently deep regressor is crucial for achieving optimal performance under the \textit{LINEAR} and \textit{MLP-3} evaluation protocols.} This finding can be explained by the disparity between transfer learning protocols: a deeper regressor is capable of generating mask representations at a higher level of abstraction, which leads to the learning of more inductive bias during fine-tuning with all parameters. Consequently, a deeper regressor performs best when fine-tuning with limited parameters, such as those in the \textit{LINEAR} and \textit{MLP-3} protocols.

\noindent \textbf{Decoder Depth.} In Table \ref{ablation} (b), we present the results of our experiments on ScanObjectNN using Point-RAE with different decoder depths. We evaluate the model's performance under three transfer learning protocols. 
\uline{The results show that the performance of the model is not significantly influenced by the decoder's depth.}
One possible explanation is that the previous MAE-based method will first implicitly predict the masked patch in the decoder, and then input it into the reconstruction head, while our proposed regressor explicit explicitly predicts the mask patch.
We find that the decoder with a 2-layer block achieves the highest accuracy. 
It is important to note that when the decoder depth is set to 0, we directly use the representation predicted by the mask regressor to achieve the reconstruction task, which is structurally different from previous works \cite{devlin2018bert, yu2022point}. 
However, we also observe that not including a decoder leads to poor results, which is consistent with previous research.


\begin{table}
\caption{
 Ablation experiments for effective of fine-tuning paradigm on the most challenging ScanObjectNN PB-T50-RS benchmark. 
 The best performances are in \textcolor{blue}{blue}.
}
\centering
\setlength{\tabcolsep}{5.5mm}
\begin{tabular}{@{}cccc}
\toprule[1pt]
 Model & FULL  & LINEAR & MLP-3 \\
\midrule[0.5pt]
(a) &89.58  &78.26  &82.96\\
(b) & \textcolor{blue}{\bf90.28} & 78.38& 83.10\\
(c) &89.07& \textcolor{blue}{\bf78.55}& 83.06\\
 (d) &89.89& 78.53 & \textcolor{blue}{\bf83.16}\\
\bottomrule[1pt]

\end{tabular}
\label{tab:finetune}
\end{table}

\noindent \textbf{Regress \& Construct.} 
To verify the effectiveness of the two major target regression and construction, we report the ablation results in Table \ref{ablation} (c). 
When both the decoder and regressor are removed, the masked modeling architecture is similar to BERT \cite{devlin2018bert}, where the encoder sees all tokens, including masked ones, and this leads to poor results.
When only the decoder is present and the regressor is removed, our model becomes similar to ACT \cite{dong2022autoencoders} and shows comparable performance in the three transfer learning protocols. However, due to incomplete functional decoupling between the encoder and decoder, the limitations of the encoder's representation learning ability still exist.
When only the regressor is present and the decoder is removed, the model directly uses the representation predicted by the mask regressor to reconstruct the target, leading to better results than previous methods and demonstrating the effectiveness of the regressor.
Finally, when both the regressor and decoder are present, our model achieves the best performance, \uline{further demonstrating regress and construct are critical for self-representation learning.}


\noindent \textbf{Mask Regressor Benefit to Both Pre-train and Fine-tune.}
We study the effect of the mask regressor for pre-train and fine-tuning for ScanObjectNN PB-T50-RS benchmark.
The results are shown in Figure \ref{loss}.
It can be seen that the reconstruct loss of the model w/ mask regressor is consistently lower than the model w/o mask regressor, which converges to a lower value more stably (0.053 vs. 0.060), indicating that \uline{the mask regressor brings superior generalization performance of the pretraining construction task. }
The mask regressor decoupled the encoder and decoder, improving the ability of the encoder to learn generalization representation and alleviate the over-fitting issue during pre-train.
The efficacy of the regressor is further demonstrated by the fine-tuning accuracy, where we observe that fine-tuning the model pre-trained with the regressor to the downstream tasks results in improved performance.

\begin{figure}[t]
  \centering
  \includegraphics[width=0.85\linewidth]{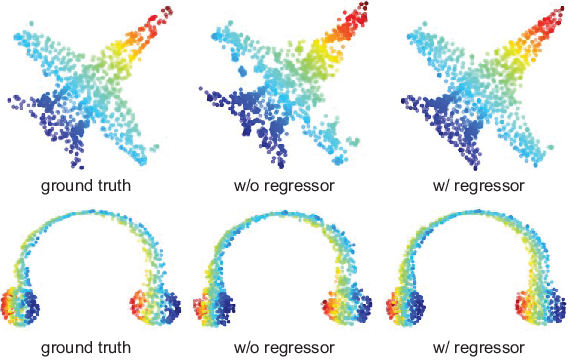}
  \caption{ Visualization of reconstruction results of synthetic objects from ShapeNet test set.}
  \label{visualization}
\end{figure}

\noindent \textbf{Fine-tune with fully using the knowledge of pre-training.}
We propose a novel fine-tuning paradigm for our Point-RAE and illustrate the different variants in Figure \ref{finetune}. The vanilla fine-tuning pipeline in Figure \ref{finetune} (a) is commonly employed by existing MAE-based methods to handle downstream tasks. However, our proposed pipeline consists of three different approaches that can be flexibly chosen for different tasks.

As shown in Table \ref{tab:finetune}, all three fine-tune pipeline are outperform the vanilla fine-tuning pipeline.
And these pipelines have slightly different effects under different fine-tune settings.
\uline{The model (b) achieves the best performance under FULL tuning protocol.}
A possible explanation is that because the mask regressor can predict the ability of the point cloud representation, the mask regressor can better predict the representation of the key points of the point cloud under FULL tuning protocol, and there is a gap between the pre-training data set and the downstream data set, so The effect is not very good under LINEAR and MLP-3 tuning protocol.
\uline{The model (c) and (d) achieve better results under LINEAR and MLP-3 tuning protocol than FULL tuning protocol.}
These models utilize the prediction ability of mask regressor to predict more representation of point clouds which do not exist in the original data.
Doing so can improve the shortcomings of point cloud sparsity, and make point cloud features denser by predicting more point representations.
Therefore, this paradigm can achieve better results with fine-tuning a small number of parameters.

\noindent \textbf{Construct Visualization.}
Figure \ref{visualization} presents a comparison of the reconstruction results obtained by models with and without the mask regressor. The results indicate that the model with the mask regressor can reconstruct high-quality object details. While both models can well reconstruct simple attributes, such as object shapes, the model with the mask regressor can better reconstruct objects with complicated details, such as the airplane wing in the first row, retaining the detailed local geometric information. 
This result can attest to the mask regressor's prediction ability for masked patches and improve the point cloud reconstruction effect, which is consistent with the conclusion of Figure \ref{loss}, and further proves the effectiveness of our proposed Point-RAE method.

\section{Conclusion}

In conclusion, the Point-RAE proposed in this paper is a novel point cloud pre-training method that effectively learns the representation of point clouds for downstream tasks. The Point-RAE employs a masked auto-encoder architecture with a mask regressor to predict the representation of masked patches, which improves the expressive ability of the learned feature space. Additionally, the proposed fine-tuning paradigm further enhances the effectiveness of the pre-trained model for downstream tasks. Experimental results on various benchmarks demonstrate that Point-RAE outperforms existing methods on different tasks.

\section{Acknowledgement}
This work was supported by the National Natural Science Foundation of China (No. 62203476).

\normalem   
\bibliographystyle{ACM-Reference-Format}
\balance
\bibliography{main.bbl}

\newpage

\appendix

\begin{table}
\caption{
 Part segmentation on ShapeNetPart dataset \cite{yi2016scalable}. 
 We report the mIoU over all classes (Cls.) and the mIoU over all instances (Inst.). The best performances are in \textcolor{blue}{blue}.
}
\centering
\begin{tabular}{@{}lccc}
\toprule[1pt]
Method &Year&  Cls.mIoU (\%)  &  Inst.mIoU (\%)
\\
\midrule[0.5pt]
\multicolumn{4}{l}{
\hspace{-0.5em}\textbf{Supervised Learning Only}}
\\
PointNet \cite{qi2017pointnet} &2016&80.39& 83.70 \\
PointNet++ \cite{qi2017pointnet++} &2017&81.85& 85.10\\
DGCNN \cite{wang2019dynamic} &2019&82.33& 85.20 \\
PointMLP \cite{ma2022rethinking} &2022&84.60 &86.10   \\
\midrule[0.5pt]
\multicolumn{4}{l}{\hspace{-0.5em}\textbf{with Self-Supervised Representation Learning (FULL)}}
\\
Transformer \cite{vaswani2017attention} &2017&83.42&85.10 \\
Point-BERT \cite{yu2022point} &2022&84.11  &85.60  \\
Point-MAE \cite{pang2022masked}&2022 &- &86.10\\
ACT \cite{dong2022autoencoders}&2023 &84.66 &86.14  \\
\textbf{Ours:Point-RAE}  &&\textcolor{blue}{\bf 84.71} &\textcolor{blue}{\bf 86.28} \\
\bottomrule[1pt]

\end{tabular}
\label{part}
\end{table}

\section{Additional Implementation Details}

In this section, we present the detailed model configuration and training settings for pre-training and fine-tuning on downstream tasks. All experiments are conducted on a single Tesla V100 GPU.

In our Point-RAE, for different resolutions of the input point cloud, we divide them into different numbers of patches with a linear scaling.
A typical input with $p=1024$ points is divided into $n=64$ point patches.
For the KNN algorithm, we set $k=32$ to keep the number of points in each patch constant.
In the backbone, the encoder and the decoder is consist of Standard Transformer with self-attention, where the encoder has 12 blocks while the decoder has 2 blocks.
And the mask regressor is consist of Transformer with cross-attention, where the mask regressor has 4 blocks.
Each block has 384 hidden dimensions and 6 heads. 
For downstream tasks, the decoder is discarded.

\subsection{Pre-training}
We use ShapeNetCore from ShapeNet \cite{chang2015shapenet} as the pretraining dataset. 
ShapeNet is a clean set of 3D CAD object models with rich annotations, including ~51K unique 3D models from 55 common object categories.
We split the dataset into a training set and a validation set but only conduct pre-training on the training set. For each instance, we sample 1024 points via FPS as input point cloud. Note that we only apply standard random \textit{Scale\&Translate} for data augmentation during pre-training.

\subsection{Classification}
For classification task, we fine-tune both the encoder and the mask regressor.
The representation of the encoder is feed into the mask regressor, then taking the output of regressor, we adopt a max pooling \& mean pooling operation and concatenate the resulted feature of two pooling.
Then, the concatenated feature is fed to the classification head.
For the \textit{FULL} tuning protocol, the classification head consist of a MLP. BatchNorm, RELU activation, and Dropout with a ratio of 0.5 are adopted in each layer of MLP.
For the \textit{LINEAR} and \textit{MLP-3} tuning protocols, the classification head are a fully connection layer and a MLP respectively, and we only tune the classification head.
We apply standard random \textit{Scale\&Translate} as data augmentation for ModelNet40 while adopt random \textit{Rotate} for ScanObjectNN.
Moreover, we use RSMix \cite{lee2021regularization} in addition to random \textit{Rotation} as data augmentation for ScanObjectNN.

\subsection{Few-shot}
For few-shot learning, we conduct the evaluation on the ModelNet40 \cite{wu20153d} dataset.
We fine-tune both the encoder and the mask regressor.
The representation of the encoder is feed into the mask regressor, then we adopt a max pooling \& mean pooling for the output of representation of both  the encoder and the regressor, and add them as the input of the classification head.

\subsection{3D Part Segmentation}
Same to classification task, we fine-tune both the encoder and the mask regressor,  the representation of the encoder is feed into the mask regressor, then taking the output of regressor, we adopt a max pooling \& mean pooling operation and concatenate the resulted feature of two pooling.
Then, the concatenated feature is fed to the segmentation head.

\section{Additional Experiments}

\begin{table}[]
    \caption{Ablation study of masking ratio. \textit{Random mask with a high ratio works best.}}
    \centering
        \begin{tabular}{cccc}
        \toprule
        Ratio & FULL & LINEAR & MLP-3\\
        \midrule
         20\% &89.34 & 77.65 & 82.34\\
         40\% &89.21 & 77.81&82.26\\
         60\% &89.78 &78.11 & 82.76\\
         80\% &\textbf{90.28} & \textbf{78.55} & \textbf{83.16}\\
         \bottomrule
        \end{tabular}
        \label{maskratio}
\end{table}

\noindent \textbf{3D Part Segmentation}
To evaluate the geometric understanding performance within objects, we conduct the part segmentation experiment on ShapeNetPart \cite{yi2016scalable}. 
The synthetic ShapeNetPart is selected from ShapeNet with 16 object categories and 50 part categories, which contains 14,007 and 2,874 samples for training and validation. 
For fair comparison, we utilize the same segmentation head as previous works \cite{pang2022masked}.
From Table \ref{part}, it can be observed that our Point-RAE improves the from scratch baseline by +1.29\% and +1.18\% Cls. mIoU and Inst. mIoU, respectively. 
This demonstrates that the mask regressor also benefits the understanding for fine-grained point-wise 3D patterns.

\begin{table}[]
   \centering
   \caption{Alignment Target. \textit{Negative cosine similarity as the alignment loss performs best.}}
    \begin{tabular}{cccc}
        \toprule
        Target & FULL & LINEAR & MLP-3\\
        \midrule
         NT-Xent loss   &89.01& 76.81 & 82.08\\
         Info NCE loss  &89.21 &77.28 & 82.16\\
         Mean Square Error &88.34& 76.31 & 81.52\\
         Negative cosine similarity &\textbf{90.28} & \textbf{78.55} & \textbf{83.16}\\
         \bottomrule
    \end{tabular}
    \label{alignloss}
\end{table}

\section{Additional Ablation Studies}
To verify the effectiveness of each component of our Point-RAE, we conduct more ablation studies on the three variant ScanObjectNN benchmarks. 

\noindent \textbf{Masking Ratio.}
In Table \ref{maskratio}, we present the results of an ablation study where we compare the effects of different masking ratios used for pre-training our Point-RAE model. Our default mask strategy follows the random masking approach proposed in \cite{pang2022masked, dong2022autoencoders}, where we randomly sample patches without replacement from the input point cloud data following a uniform distribution. We set the default masking ratio to a high value of 80\% for our experiments.
The results in Table \ref{maskratio} demonstrate that using a higher masking ratio with random masking leads to better performance for all three variants of transfer learning protocols.

\noindent \textbf{Alignment Target.}
The alignment target is a crucial factor in predicting the representation of masked patches, and it has a significant impact on the downstream task results.
In Table \ref{alignloss}, we compare the performance of different alignment targets.
As our proposed Point-RAE uses cosine similarity to compute the reconstruction loss, we find that the negative cosine similarity achieves the best results.
Using negative cosine similarity helps to maintain the consistency between the prediction and reconstruction feature spaces during training, which is beneficial for improving the expressive ability of the feature space.
 \end{document}